# Passive Luminescent Bellows Mechanism

Naoto Kikuta, Issei Onda, Kazuki Abe, Masahiro Watanabe, Kenjiro Tadakuma

*Graduate School of Engineering Science, Osaka University*
*560-8531, D442, 1-3 Machikaneyamacho, Toyonaka-shi, Osaka, Japan*
naoto.kikuta@pumech.sys.es.osaka-u.ac.jp
kenjiro.tadakuma@pumech.sys.es.osaka-u.ac.jp

*Abstract*—The use of robots in disaster sites has rapidly expanded, with soft robots attracting particular interest due to their flexibility and adaptability. They can navigate through narrow spaces and debris, facilitating efficient and safe operations. However, low visibility in such environments remains a challenge. This study aims to enhance the visibility of soft robots by developing and evaluating a passive luminescent flexible actuator activated by a black light. Using Ecoflex mixed with phosphorescent powder, we fabricated an actuator and confirmed its fluorescence, phosphorescence, and deformation ability. Furthermore, the effects of the mixing ratio on optical and mechanical properties were assessed.

## I. Introduction

In recent years, the use of robots in rescue operations at disaster sites has expanded rapidly. In the event of a disaster, robots are able to undertake tasks that are difficult for humans, such as searching for victims under rubble and assessing the conditions inside buildings—tasks that are challenging to achieve with traditional methods [1, 2]. Furthermore, the introduction of robotic technology has greatly improved the efficiency of tasks such as debris removal, obstacle removal, and equipment installation [3, 4]. Among the various types of robots, soft robots, which can adapt to confined spaces through shape transformation, are expected to play a significant role in future rescue operations at disaster sites. This is because of their potential to access areas that are otherwise physically unreachable [5, 6].

Soft robots, due to their flexibility and adaptive motion, contribute to enhancing the efficiency and safety of rescue operations at disaster sites. Their flexible structures enable them to perform tasks that are difficult for conventional rigid robots, such as searching for victims in narrow spaces and assessing the situation [7]. Additionally, by utilizing their flexibility and lightweight characteristics, they can absorb impacts during collisions, minimizing damage from contact with obstacles or humans, thereby ensuring safe operations [8]. Given these advantages, soft robots are expected to be a promising technology for disaster rescue operations.

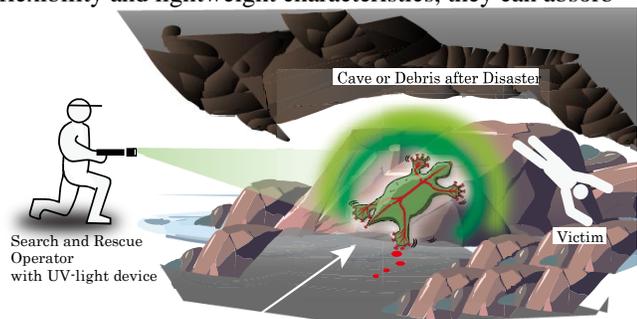

**Fig.1-1:** Concept of the Soft Search and Rescue Robot with Passive Luminescent Elements

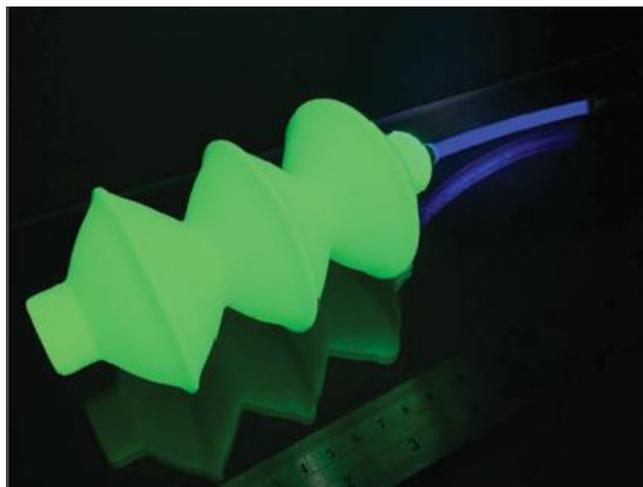

**Fig.1-2:** Overview of the First Prototype Model of Passive Luminescent Bellows Mechanism

However, soft robots face several challenges before they can be practically applied, with visibility in low-light environments being one of the most prominent issues. This challenge becomes particularly severe when working under rubble or at night. During disasters, bright lighting is not always available, and visibility is often obstructed by power outages or debris, making it extremely difficult to track the movement and position of robots that rely on visual information. For example, when searching for victims trapped under rubble or assessing the conditions inside damaged buildings, robots often need to operate in darkness. In such cases, inadequate visibility makes it difficult to accurately track their position, leading to a significant decrease in the efficiency and safety of rescue operations, as well as potentially hindering collaboration with other rescue teams. Therefore, improving visibility in low-light environments is a critical issue that needs to be addressed to ensure the accurate operation and safe rescue activities of soft robots.

To address this visibility issue, making the robots themselves emit light is considered to be an effective solution. Specifically, by incorporating a lighting function into the actuators that make up the robots, their position and movement can be visually tracked in low-light environments at disaster sites, leading to improved operational efficiency and safety. Moreover, such lighting can facilitate collision avoidance between robots and enable smooth collaboration with human operators.

On the other hand, realizing this lighting function also presents several challenges. First, implementing a lighting function requires the installation of large batteries, which may compromise the greatest advantages of soft robots—flexibility, lightweight, and compact size. Additionally, since lighting consumes energy, this can be a major drawback in disaster sites where long operating times are required.

To address these challenges, passive lighting is considered an effective approach. By using passive lighting, it may be possible to enhance visibility in low-light environments while mitigating issues such as the need for large batteries or a significant increase in the size of the robot.

The purpose of this research is to develop and materialize the concept of a passive-lighting flexible actuator. For this, we will develop and evaluate materials for an actuator that glows brightly when illuminated with black light. The actuator will feature a bellows structure and use gas as the working fluid. The goal is to enhance visibility while maintaining flexibility and strength, ensuring that the robot performs reliably under actual operational conditions. Figure 1 shows the fluorescent and phosphorescent properties of the actuator realized in this study.

## II. FUNDAMENTAL PRINCIPLE

2.1 Mechanism of Passive Light Emission Using Ultraviolet Rays

A black light is a light source that primarily emits ultraviolet (UV) radiation with wavelengths below 400 nm. While much of this UV radiation is invisible to the human eye, when it irradiates fluorescent or phosphorescent materials, the materials absorb the ultraviolet energy and re-emit it as visible light. Fluorescent materials emit visible light almost instantly, while phosphorescent materials gradually release the absorbed energy, allowing them to continue glowing even after the light source has been removed.

2.2 Mechanism of Fluorescence and Phosphorescence

The difference between fluorescence and phosphorescence is explained below.

Fluorescence occurs when a material absorbs energy from an external light source, causing its electrons to transition to an excited state. As the electrons return to their ground state in a very short period, they release the absorbed energy in the form of visible light [9]. A key characteristic of fluorescence is that the energy is emitted immediately; therefore, the light stops as soon as the light source is removed.

On the other hand, phosphorescence occurs when the absorption of light causes the electrons in the material to become excited and transition to higher-energy outer orbitals. Once the absorption ceases, the electrons gradually return to their original lower orbitals over time, releasing the stored energy as visible light during this process [10]. This emission can last from several minutes to several hours, providing visibility in dark environments even after the light source has been removed.

2.3 Basic Units of Light and Measurement Methods

In light measurement, two basic units are commonly used: lumens (lm) and lux (lx). Their definitions are as follows:

Lumen (lm) is the unit that represents the total amount of light emitted by a light source. It indicates the intensity of the energy radiated by the light source and is also referred to as luminous flux.

Lux (lx), on the other hand, is the unit of illuminance, representing the amount of light falling on a specific area. The illuminance
I (in lx) is calculated using the following equation:
The illuminance
I (in lux) is given by the following formula:

$$I = \frac{P}{4\pi r^2} \quad \ldots\ldots\ldots\ldots (1)$$

Where:
$I$: I is the illuminance at the target location (in lx),
$P$: P is the total luminous flux emitted by the light source (in lm),
$r$: r is the distance between the light source and the target surface, and represents the surface area of a sphere over which the light energy is uniformly spread.

According to this equation, as the distance r from the light source doubles, the illuminance I decreases in proportion to the square of the distance, reducing to 1/4 of its original value. This inverse-square law forms the basis for understanding how distance affects illuminance in lighting design.

## III. PROTOTYPE IMPLEMENTATION

In this experiment, the proposed flexible actuator has two key characteristics: strong luminescence when exposed to blacklight and the ability to store and release light (phosphorescence), as well as excellent performance as a pneumatic actuator. To achieve these features, we determined that silicone rubber with high flexibility and tensile elongation, combined with luminescent materials to enhance the luminescence and phosphorescent capabilities, would be optimal. Specifically, Ecoflex 00-30 (Smooth-On Inc., USA) was selected as the material for the proposed actuator due to its superior flexibility and suitability for mixing with luminescent materials.
A bellows actuator, as shown in Figure 2, was realized using this material. The mixing ratio of liquid Ecoflex to phosphorescent powder used was 5ml:1g.

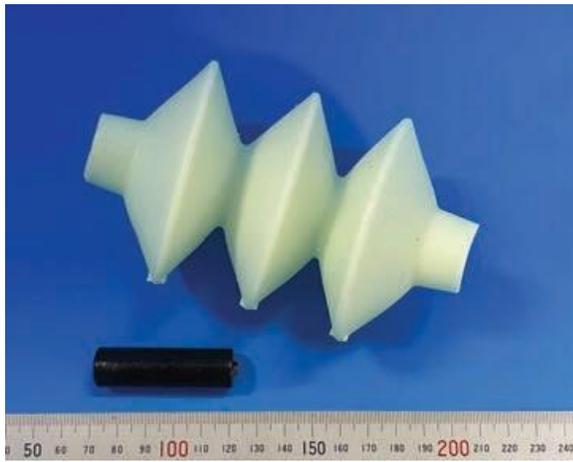

**Fig. 2: Appearance of the fabricated prototype actuator**

## IV. BASIC EXPERIMENTS

In this experiment, the luminescence of the Ecoflex and phosphorescent powder mixture, which forms the material of the bellows, was first investigated to determine whether it glows. Following this, changes in the light intensity and mechanical properties were studied based on the amount of phosphorescent powder mixed in.

4.1 Evaluation of the Luminescence and Expansion/Contraction Performance of Phosphorescent Powder-Mixed Bellows

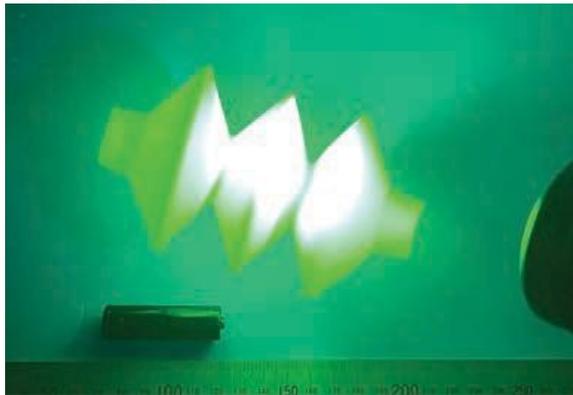

**Fig. 3: Fluorescence behavior under black light**

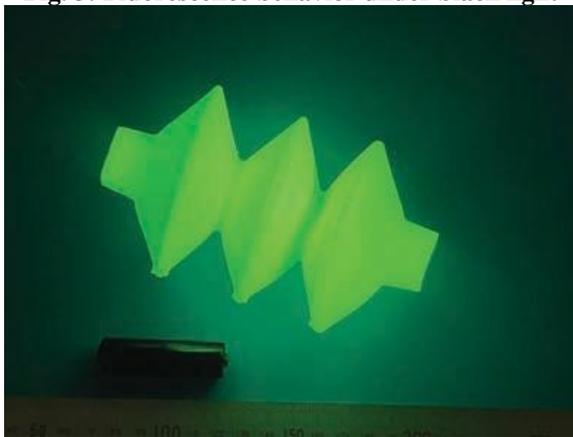

**Fig. 4: Phosphorescence behavior after black light exposure**

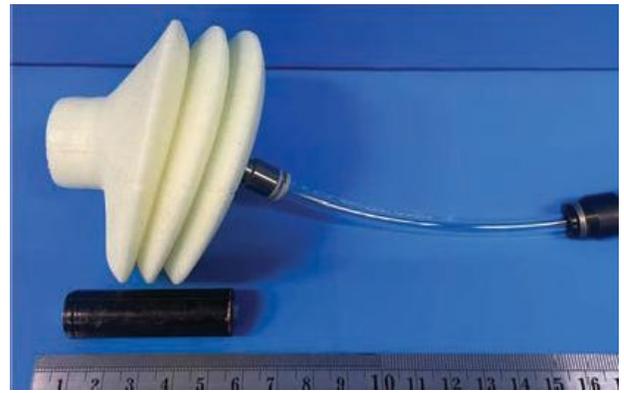

**Fig. 5: Contracted state of the actuator**

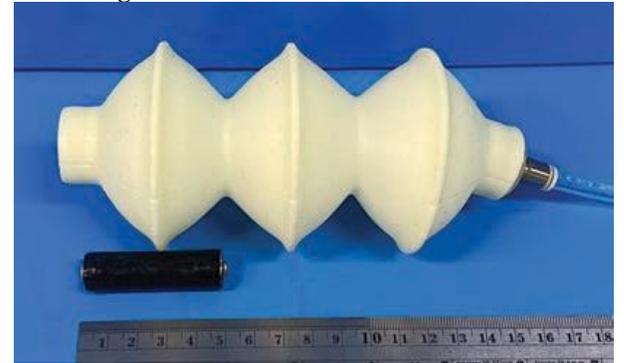

**Fig. 6: Expanded state of the actuator**

The bellows actuator realized in Chapter 3 was evaluated for its fluorescence performance under blacklight, its phosphorescence performance after the blacklight was turned off, and its expansion/contraction performance when positive and negative pressures were applied. As shown in Figures 3 to 6, all performance aspects—fluorescence, phosphorescence, contraction, and expansion—were confirmed. Therefore, this actuator is considered to meet the requirements as a passive luminescent actuator.

4.2 Light Intensity Changes Due to Mixing Ratio

An experiment was conducted under the conditions shown in Figure 7 to observe how the intensity of light (measured in lux) varies according to the mixing ratio of phosphorescent powder. The procedure is outlined as follows, and for each test sample, phosphorescent powder was mixed into 50 ml of Ecoflex:

Black light was applied to the test sample for 10 seconds, and the illuminance was measured every second during this period.

The average lux value over 10 seconds was recorded as the illuminance of the test sample.

After stopping the blacklight, the change in illuminance due to phosphorescence was measured for 1 minute.

Steps 1 to 3 were repeated for different test samples, each with varying amounts of phosphorescent powder mixed into 50 ml of Ecoflex.

The experimental results are shown in Figures 8 and 9. The initial illuminance (lux) was defined as the illuminance at 11 seconds (the start of phosphorescence), and the duration of phosphorescence (s) was defined as the time from the start of phosphorescence until the illuminance dropped to 0 lux.

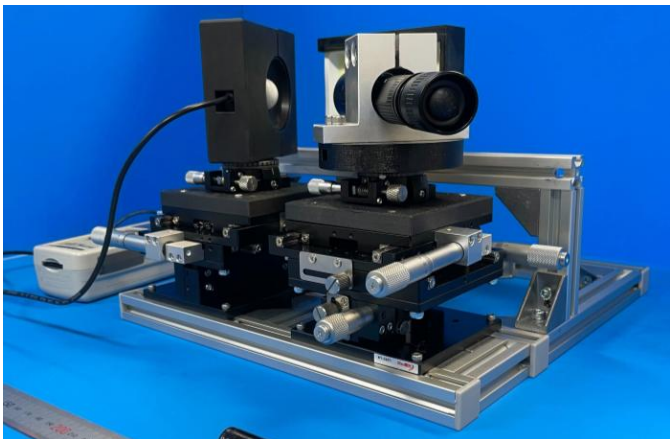

**Fig. 7a: Experimental setup and procedure for illuminance measurement**

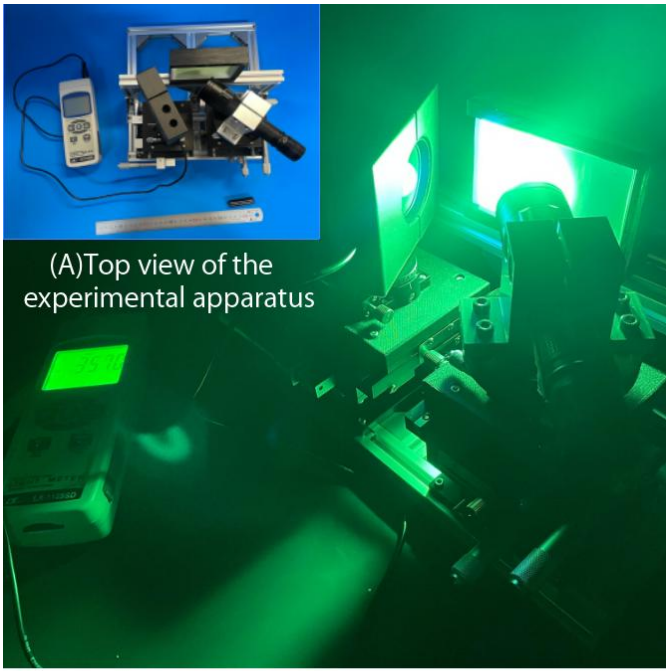

**Fig.7b: Experimental setup and procedure for illuminance measurement**

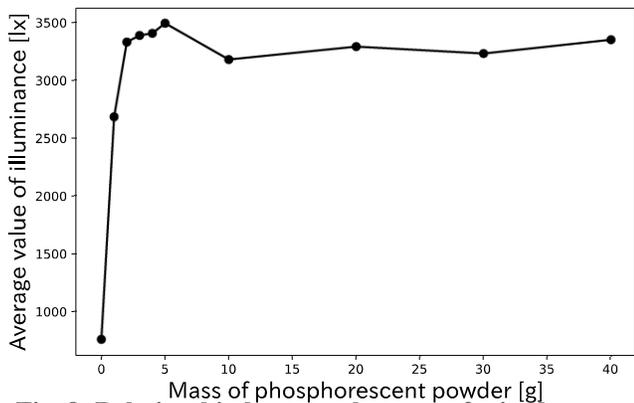

**Fig. 8: Relationship between the mass of mixed phosphorescent powder [g] and average illuminance from fluorescence [lx]**

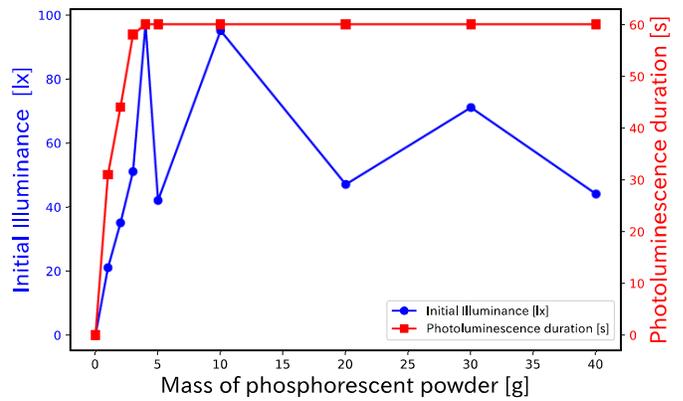

**Fig. 9: Relationship between the mass of mixed phosphorescent powder [g], initial illuminance from phosphorescence [lx], and phosphorescence duration [s]**

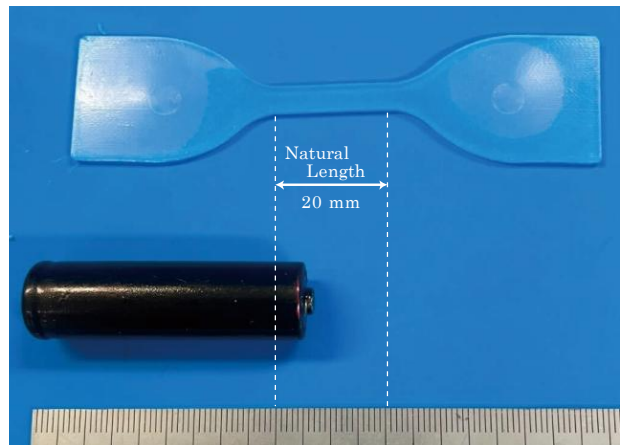

(a)  Overview of the Sample of Tensile Piece

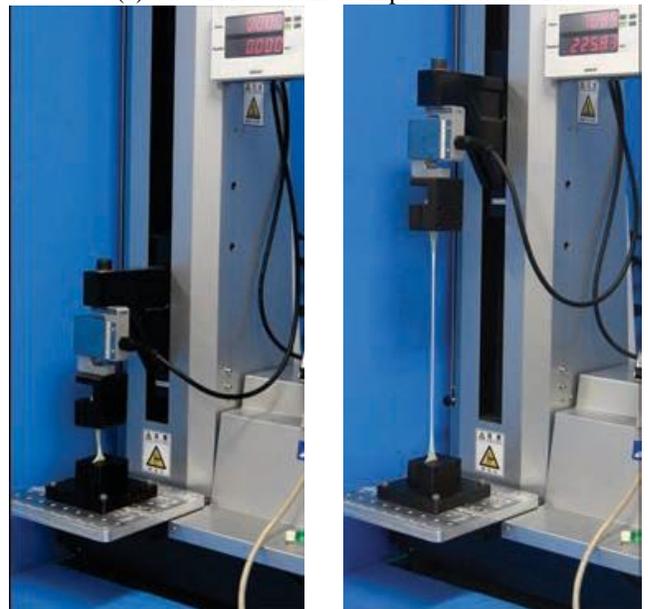

(b-i) Initial state    (b-ii) Just before fracture

**Fig. 10: Tensile test procedure**

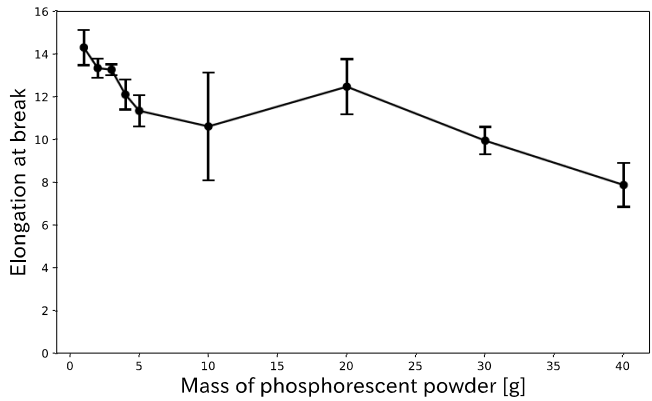

Fig. 11: Relationship between the mass of mixed phosphorescent powder [g] and elongation at fracture

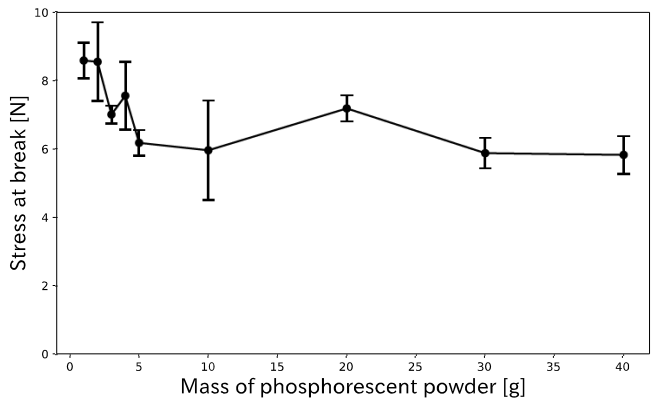

Fig. 12: Relationship between the mass of mixed phosphorescent powder [g] and stress at fracture [N]

From Figure 8, it can be seen that the luminescence performance under black light increased as the amount of phosphorescent powder increased from 0 g to 2 g, but plateaued beyond 2 g. Regarding the phosphorescent performance after stopping the blacklight, Figure 9 shows that the initial illuminance and phosphorescence duration increased as the powder increased from 0 g to 3 g. However, beyond that, no consistent pattern was observed in the initial illuminance. This irregularity may have been caused by the sampling interval of the illuminance meter (1 second), making it difficult to accurately capture the maximum illuminance at the moment the blacklight was turned off.

4.3 Mechanical Property Changes Due to Mixing Ratio

Using a tensile testing machine, the elongation and stress of the test samples used in section 4.2 were measured. The test samples followed the JIS K 6251 standard, with dumbbell-shaped Type 3 specimens. A mold was used to shape the dumbbell specimens, and a vacuum degassing process was employed to prevent air bubbles from forming in the silicone during molding. The conditioning method adhered to the JIS K 6250 standard. Figure 10 shows the tensile testing process.

In this experiment, the natural length of the test sample was defined as 0 displacement, and the displacement and stress at the time of fracture were measured. The point of maximum stress was defined as the fracture point (after which stress drops sharply). Each test sample underwent three repetitions.

From Figure 11, it was confirmed that as the amount of phosphorescent powder increased, the elongation tended to decrease, falling by approximately 50% from 0 g to 40 g. However, there was some variation, such as the elongation at 20 g being greater than at 5 g and 10 g.

As shown in Figure 12, the stress decreased by approximately 70% as the phosphorescent powder increased from 0 g to 5 g. Beyond 5 g, no significant changes were observed, though there was no change between 1 g and 2 g, and a temporary increase was observed between 3 g and 4 g. Additionally, the stress was unusually high at 20 g, indicating variability. This variation could be due to the use of molds for making the test samples and the thinness of the samples, which may have led to the formation of air bubbles or cuts during the manufacturing process. Since the number of trials was only three, increasing the number of trials in the future is expected to improve the accuracy of the data.

## V. FUTURE DEVELOPMENTS

The fundamental principle of passive luminescence in the mechanical element introduced here is not limited to actuators; it can also be applied to sensors, processors, batteries, and other components. From a materials perspective, we have confirmed that resins, vitamin B2, and fluorescent brighteners found in printing paper exhibit strong luminescence after exposure to black light. We plan to continue exploring and realizing these principles in practical applications.

In particular, we are considering replacing the working fluid of robots with a vitamin B2 solution. This luminous liquid would serve as the internal fluid of the robot, and in the event of a malfunction, it would leak out and passively illuminate, enabling early detection of the fault. This innovation aims to make robot condition monitoring easier, even in harsh environments like disaster sites, where rapid recognition of failures is crucial—something difficult to achieve with conventional technologies. Additionally, we are also exploring the application of origami structures utilizing printing paper containing fluorescent brighteners.

## VI. CONCLUSION

In this study, we proposed the principle and conducted the prototyping and material evaluation of a flexible actuator exhibiting strong fluorescence and phosphorescence under blacklight exposure. Specifically, a bellows-type actuator was fabricated using a mixture of Ecoflex and phosphorescent powder. As a result, the actuator was confirmed to demonstrate fluorescence, phosphorescence, contraction, and expansion characteristics. Additionally, performance evaluations were conducted on test samples with varying mixing ratios of phosphorescent powder, and data related to optical intensity and mechanical properties were collected.

As future work, the limitations related to the resolution and sampling interval of the illuminance meter should be addressed, as they prevented accurate measurement of the initial illuminance and the exact onset of phosphorescence. More precise instrumentation is needed to obtain accurate measurements. Furthermore, given the small number of trials in the current study, which led to several outliers, increasing the number of trials is essential to improve the reliability and statistical validity of the data.